\documentclass[aps,preprint]{revtex4}
\usepackage{amsfonts}
\usepackage{amsmath}
\usepackage{amssymb}
\usepackage{graphicx}

\begin{document}
\preprint{ }
\title[Short title for running header]{Size dependent word frequencies and translational invariance of books}
\author{Sebastian Bernhardsson, Luis Enrique Correa da Rocha, and Petter Minnhagen}
\affiliation{Dept. of Physics, Ume\aa \ University. 901 87 Ume\aa . Sweden}

\begin{abstract}
It is shown that a real novel shares many characteristic features with a null model in which the words are randomly distributed throughout the text. Such a common feature is a certain translational invariance of the text. Another is that the functional form of the word-frequency distribution of a novel depends on the length of the text in the same way as the null model.
This means that an approximate power-law tail ascribed to the data will have an exponent which changes with the size of the text-section which is analyzed.
A further consequence is that a novel cannot be described by text-evolution models like the Simon model.
The size-transformation of a novel is found to be well described by a specific Random Book Transformation.
This size transformation in addition enables a more precise determination of the functional form of the word-frequency distribution.
The implications of the results are discussed. 

\end{abstract}

\maketitle

\section{Introduction}

Some 75 years ago Zipf found that the word frequency of a language has a very
particular ``power-law like'' distribution
\cite{Zipf32}. This phenomena is best known as Zipf's law and states that the
number of occurrences of a word in a long enough written text falls off as
$1/r$ where $r$ is the occurrence-rank of a word (the smaller rank, the more
occurrences) \cite{Zipf32} \cite{Zipf35} \cite{Zipf49} \cite{mitzenmacher03} \cite{baayen01}.
How well is this power law obeyed? What is its origin? What does it imply from
a linguistic and cognitive point of view, if anything?

Simon in Ref.\ \cite{simon55} emphasized that the fact that ``power law''
distributions occur in a wide range of seemingly
unrelated phenomena suggests a general underlying stochastic nature. In
particular he devised a general stochastic model for the writing of a text,
the \emph{Simon model} \cite{simon55}. The random element in this model is
tied to the actual process of evolving the text and not to a property of the
language itself. The \emph{Simon model} and its stochastic evolution mechanism
has since its first appearance turned up in many disguises such as
rich-get-richer models and preferential attachment \cite{newman05}. An
alternative view was taken by Mandelbrot who proposed that Zipf's law of word
frequencies could be associated with the collective language itself rather
than with the evolution of a particular text \cite{mandelbrot53}. In
particular he proposed that the ``power-law
like'' distribution could be linked to an optimization of a
letter-combination information \cite{mandelbrot53}. However, Miller in
Ref.\ \cite{miller57} showed that a power law distribution of words in a
collective language does not \emph{per se} requer any optimization, which gave
rise to the metaphor of a monkey randomly writing on a typewriter \cite{li92}.
All these proposed explanations presumes that the
``power-law like'' distribution says nothing about the syntax,
grammar and context correlations of a written text. Yet the word correlations
are, of course, essential for the meaning of a text.

In the present paper we focus on the function $W_{D}(k)$, the number of
distinct words which occur precisely $k$ times in a written text. The
correspondence of Zipf's word rank power law is for this quantity
$W_{D}(k)\sim1/k^{2}$ \cite{simon55}. We here focus on the properties of single novels, each novel written by a single author.
In this way we ensure that both the evolution aspect of
the text and the properties of the language always relates to the very same text.
From this perspective a novel can perhaps be regarded as a fingerprint of the
author's brain \cite{gonzalves06}. We demonstrate that the text of a novel display certain general features and show that these features are shared with a simple null model which we call the random book.
 
In section 2 we describe some general characteristic features which the text of a novel display. For clarity we choose one typical novel as an illustrative example. In appendix A we include data for a collection of novels in order to illustrate the generality of the conclusions. In section 3 we discuss the random book transformation which describes how the word-frequency distribution changes with the length of the text analyzed. It is shown that a real novel to good approximation  transforms in the same way. It is also shown that the random book transformation can be used in order to obtain a sharper determination of the word-frequency distribution of a novel. Section 4 contains our summary and concluding remarks.

\section{Bookish Facts}

Examples of key characteristics of the word frequencies in a novel are as
follows: The most obvious is the word-frequency distribution of the
\emph{complete} novel. A word is in this context defined as a group of letters
separated by blanks. If the book contains $W_{D}$\ distinct (different) words
and a total of $W_{T}$\ words, then $P(k)=W_{D}(k)/W_{D}$ is the probability
that a word, which you pick randomly in the book, is occurring $k$-times in the book. This means that $\sum
_{k=1}^{k_{\max}}P(k)=1$ where $k_{\max}$\ is the maximum number of times a
distinct word appears in the book and also that $\sum_{k=1}^{k_{\max}%
}kP(k)=W_{T}/W_{D}=\langle k\rangle$, which is the average number of times a
word occurs in the book. The function $P(k)$ is the word-frequency
distribution and is often very broad and more or less ``
power-law like'', $P(k)\sim1/k^{\gamma}$ with $\gamma\leq2$,
over a substantial region. This is illustrated in Fig.\ 1a with data for the
novel \emph{Howards End} (HE in the following) by E.\ M.\ Forster taken from
Ref.\ \cite{gutenberg} where circles correspond to the raw data.
The horizontal distribution for the largest $k$-values means that only
single unique words have the largest number of occurrences. The triangles
corresponds to a $\log_2$-binning (bin $i$ has a size of $2^{i-1}$) of the data and one notes that these data follow
a smooth curve. \emph{This last fact implies that the data are produced by a
stochastic process}. The functional form $P(k)\sim\exp(-bk)/k^{\gamma}$ gives
a good fit to the data ($\gamma=1.73$ in Fig.\ 1a). The level of "goodness" of this fit is discussed
in section III and shown in Fig.\ \ref{4}.

Instead of analyzing the complete book, one can analyze a \emph{section} containing a total of $w_{T}%
$ words. Then one finds that the ``power-law
slope'' of the corresponding word-frequency distribution,
$P_{w_{T}}(k)$, in a novel depends on the total number of words $w_{T}$. This
is illustrated in Fig.\ 1b, which shows the average word-frequency distribution
for $n^{th}$-parts of Howards End. The total number of words is $W_{T}%
\approx110000$ which means that the $n=20$-part shown in Fig.\ 1b only has
$w_{T}=\frac{W_{T}}{n}\approx5500$ words while the $n=200$-part corresponds to
$w_{T}\approx550$ words. The word-frequency distribution for a section of size
$w_{T}$ is obtained as an average over a large amount of sections of the same
size and we use periodic boundary conditions in order to avoid reduced statistics
due to the boundaries of the book. As will be shown below, real books display a strong tendency
towards having the words close to randomly distributed, allowing for the use of periodic boundary conditions.
As seen in Fig.\ 1b the slope of the ``power-law like`` part of the distribution gets systematically steeper
when taking smaller and smaller sections of the book.
From a practical point of view this means that if you attempt to approximate
the word frequency distribution with the function $P_{w_T}(k)\sim
\exp(-bk)/k^{\gamma}$ then the exponent $\gamma$ increases as $w_T$ 
decreases. \emph{The change of the shape of} $P_{w_T}(k)$ \emph{as a
function of the total number of words} $w_T$ \emph{is a characteristic
feature of the word frequency in a book}.

Fig.\ 2a shows the number of distinct words $w_D(w_T)$ as a function of
the total number of words $w_T$: the first word is always distinct which
means that $w_{D}(w_{T}=1)=1$. As you go further into the book, words tend to
be repeated which means that the number of distinct words increases slower than a straight line with slope 1.
The shape of $w_{D}(w_{T})$ gives a characteristics of the novel since it reflects the spatial distribution of words within the novel.
Note that the function $w_{D}(w_{T})$ and the distributions $P_{w_{T}}
(k)$ are directly related, since the average number of times a distinct word
appears is $\langle k \rangle_{{w_{T}}}=\sum_{k=1}^{k_{\max}}kP_{w_{T}
}(k)=1/\frac{w_{D}}{w_{T}}(w_{T})$. How would $w_{D}(w_{T})$ change if the
words were completely randomly distributed in the book, keeping the same frequency distribution? As seen from Fig.\ 2a,
the function for the randomized book (where all words are placed randomly in
the book) is very close to the raw data of the novel. \emph{A characteristic feature of a novel is that the distribution $w_{D}(w_{T})$ is close to the one for the random null model of the novel}. This implies that the real novel and the null model share some overall random features. 

The random features are also reflected in the distribution of words belonging to different frequency classes: the frequency class $k$ contains all words which appears precisely $k$ times in the book. For example the class $k=1$ contains all the words which only occurs once in the book. Random with respect to frequency classes means that there is no preference for words belonging to a specific frequency class to appear in any particular part in the book. Thus for a random book you should have encountered close to half the words belonging to a frequency class when you have read precisely half of the book. Fig.\ 2b shows the percentage of words belonging to a frequency class $k$ encountered after reading half a real book as a function of $k$. The data is for the real HE and the full drawn horizontal line is the expectation value for a randomized HE.
The grey shadings mark one and two standard deviations (using the same binning as for the real novel) away from the randomized HE. This means that if the data circles in Fig.\ 2b had belonged to a single realization of a random HE book then they would with large probability fall inside the grey areas. The actual circles in Fig.\ 2b give the data for the real novel HE. These data follow the same horizontal trend and are compatible with the random null model over a substantial region of $k$ values. However a real novel is of course a highly purposely structured creation. Some noticable deviations in Fig.\ 2b can immediately be associated with such contructive features. The first noticable deviation in Fig.\ 2b is that the value for the frequency class $k=1$ (words which only occur once in the book) is only 47\% (an average over the collection of books in Appendix A gives 47,3\%), which is a statistically significant deviation from 50\%. The reason is that an author who writes a book from the beginning to the end will have a slightly decreasing tendency of introducing new rare words towards the end of the book. Another noticable deviations is the two circles higher than 50\% for larger $k$ (words occurring very often in the book). These deviations are actually caused by the two specific words \emph{she} and \emph{her} and are clearly context related features in the novel (a particular context in chapter four about a third into the book has a very low concentration of \emph{she} and \emph{her}). Nevertheless Fig.\ 2b illustrates that the overall tendency of the data has the same characteristic feature as the null model. 
For the Simon model, the distribution of words belonging to different frequency classes are incompatible with the random feature displayed by real novels: The triangles in Fig.\ 2b represent the data for a single Simon book of the same length and $\langle k \rangle$ as HE. 
The dashed lines give the analytic asymptotic behavior in the small and large $k$ limits (see Appendix B). It is clear that rare words tend to appear very late in the Simon-book while common words are more densely positioned early in the book. As explained below, this is because the Simon model is a growth model.

Another characteristic feature of the null model is that the text is translationally invariant. 
This means that if you divide the novel into three consecutive sections and obtain the functions $w_{D}(w_{T})$ separately for all three sections then these three functions show no systematic trend in there deviations. Fig.\ 2c demonstrates that the same is to very good approximation also true for the real novel HE. Appendix A gives data from a variety of novels suggesting that the qualitative agreements between the random null model and real novels given by Figs.\ 2a and c are indeed general features. Real books contain information in the form of a story. Different parts describe different events and surroundings which may creates word correlations. So, we should expect some fluctuations between curves for different parts of a novel. But the point is that, in general, no systematic change can be observed between parts of a real novel.
\emph{The translational invariance of the text is a characteristic feature of a novel.}

Whereas a real novel is in qualitative agreement with the null model, the \emph{Simon model} is instead incompatible: Fig.\ 2d shows that the Simon model does not obey translational invariance, but instead display a strong systematic trend.
The data is obtained by generating books of the same length as HE using the stochastical growth model by Simon \cite{simon55}. The books are divided into three consecutive parts of equal size and the average distributions for these three parts are plotted in the figure. As seen the distributions systematically changes with the position in the book in a way that is incompatible with the translational invariance. This is contrary to the data for a real book (compare Fig.\ 2c).
So what is wrong with the Simon model in the context of real novels? 
The problem can be traced to the stochastic element (the dice) in the model:
The ground version of the \emph{Simon model} goes as follows\cite{simon55}: The novel is
assumed to be written by adding words in a consecutive order from the start to
the end. Each time the author adds a word to the text it can either be a
word not previously used in the text or an old word. There is a certain chance
to add a new word and a certain chance to use an old one. The crucial stochastical
assumption in the model is that the chance for picking a specific old
word is directly proportional to the number of times this word has already
been used in the text written so far. Thus the randomness in the Simon model
is associated with picking words randomly from the text already written. As
this text evolves, the reservoir (the text written so far) from which the
random words are picked also changes. Hence the random element in Simon-type
models explicitly depends on the growth process of the text. It means that the stochastic element changes with the position in the book. This is in contrast to the random null model, where the randomness is independent of the position in the book.
One may also note that the resulting word-frequency distribution, $P(k)$, for the Simon model
,with a constant growth rate, is independent of the length of the text. This is in contrast to a real novel where the shape of the distribution changes with the length of the text (compare Fig.\ 1b). 
The crucial point is that stochastic text evolution models in general have the same problem as the Simon model, including all preferential
attachment type models \cite{dorog01} \cite{masucci06} \cite{barabasi99} \cite{newman06}.
\emph{Growth processes which are based on a stochastic element (a dice) which ipso facto depends on the position in the text do not adequately reproduce the statistical distribution of words in a text.}
We emphasize that this is a fundamental structural feature which cannot be remedied within this class of stochastic models. This implies that the stochastic element in real novels belong to an altogether different stochastic class.

A noteworthy additional characteristic feature is that the word-frequency
distribution $P_{w_{T}}(k)$ for an author does to large extent only depend on
the number of words $w_{T}$ written by the author and not on the specific book
or short story. This is illustrated in Fig.\ 3 by comparing a short story by
D.\ H.\ Lawrence (\emph{The Prussian Officer} (PO), $W_{T}\approx9000$) with book
sections of the corresponding size from two of his full novels. Fig.\ 3a is for \emph{Woman in Love} (WL) which has $W_{T}\approx180000$ and b) for \emph{Sons and Lovers} (SL) which has $W_{T}\approx162000$. As in the case of Fig.\ 1b, the word frequency distribution for a section is the average over many sections of the same size. In order to obtain a section size of the same length as the short story we use $n=20$-parts in a) and $n=18$-parts in b). The agreement is very good in both cases except for the data of the very highest $k$-values. This difference is an artifact of comparing a snapshot (PO) with a curve resulting from averages (sectioning of WL and SL).

\section{The Random Book Transformation}
We now return to the characteristic size dependence of the word-frequency distribution $P_{w_{T}}(k)$ for a novel described in Fig.\ 1b. In Fig.\ 4a compares this size dependence with the corresponding size dependence of the random null model: first we extract, directly from the raw data, the $P_{\omega_T}(k)$ corresponding to sections $n=200$-parts of the novel HE. This data is represented by squares in Fig.\ 4a. Next we randomize the words in HE. Note that a randomization leaves the frequency distribution $P(k)$ invariant. From a sample of the randomized HE-book we extract $P_{\omega_T}(k)$ corresponding to $n=200$-parts of the randomized HE. This is given by the triangles in Fig.\ 4a. The overlap of the data is near perfect, indicating that the null model transform in very much the same way as the real novel. 
In case of the random null model one can straighforwardly obtain the size transformation.   
The starting point is the word-frequency distribution $P(k)$ for a book with $W_{T}$ total words and $W_{D}$ different
words. The question is how $P(k)$ relates to the word-frequency distribution, $P_{w_{T}%
}(k)$, for a section size $w_{T}<W_{T}$ of the very same book. For the case
when the words within a frequency class are randomly distributed the relation follows from 
combinatorics. The probablility for a word that 
appears $k^{\prime}$ times in the full book to appear $k$ times in a
smaller section ($k^{\prime} > k$) can be expressed in  
binomial coefficients \cite{baayen01}: if we let $P(k)$ and $P_{w_{T}}(k)$ be two column 
matrices with $W_{D}$ elements numerated by $k$, then

\begin{equation}
\boldsymbol{P}_{w_{T}}(k)=C\sum_{k^{\prime}=k}^{W_{D}}\boldsymbol{A}%
_{kk^{\prime}}\boldsymbol{P}(k^{\prime})
\label{1}
\end{equation}

where $A_{kk^{\prime}}$ is the triangular matrix with the elements%

\begin{equation}
A_{kk^{\prime}}=(\frac{W_{T}}{w_{T}}-1)^{k^{\prime}-k}\frac{1}{(\frac{W_{T}%
}{w_{T}})^{k^{\prime}}}\binom{k^{\prime}}{k}
\label{2}
\end{equation}

and $\binom{k^{\prime}}{k}$ is the binomial coefficient. The coefficient
$C$ is%

\begin{equation}
C=\frac{1}{1-\sum_{k^{\prime}=1}(\frac{W_{T}-w_{T}}{W_{T}})^{k^{\prime}%
}P(k^{\prime})}
\label{3}
\end{equation}

Since $A_{kk^{\prime}}$ is a triangular matrix with only positive definite
elements it also has an inverse which is given by

\begin{equation}
A_{kk^{\prime}}^{-1}=(\frac{W_{T}}{w_{T}}-1)^{k^{\prime}-k}(\frac{W_{T}}%
{w_{T}})^{k}(-1)^{k^{\prime}+k}\binom{k^{\prime}}{k}%
\label{4}
\end{equation}

One should note that RBT (Random Book Transformation) only hinges on the assumption that words belonging to a frequency class are randomly distributed through out the book. Since this assumption is rather well obeyed by real novels (compare Fig.\ 2b), the near perfect agreement between the randomized null model and the real HE in case of the two $n=200$-parts shown in Fig.\ 4a may be interpretated as a confirmation that the real novel and the randomized novel share some basic stochastical features.  

In Fig.\ 4b we start from the randomized HE and section it into parts with $w_{T}$ words. From each section size the average number of distinct words $w_{D}$ is determined so that one obtains the quantity $1/\langle k \rangle_{w_{T}}=\frac{w_{D}}{w_{T}}(w_{T})$. An average over many sections of the same size is used. The result is the full drawn curve in Fig.\ 4b. One should note that this is in fact not a curve but a very dense set of data points (each point corresponds to a different section size which means that the total number is $W_T \approx 110000$). In this way the raw data for HE given by the cirles in Fig.\ 1a are transformed into a very smooth curve for $\frac{w_{D}}{w_{T}}(w_{T})$. The Bayesean probabalistic assumption used is that words from different word-frequency classes have no preferential order. As apparent from Fig.\ 2b and Fig.\ 4a this is a very reasonable Bayesean assumption. The point is now that the function $\frac{w_{D}}{w_{T}}(w_{T})$ through the RBT-transformation uniquely determines $P(k)$ and vice versa. In order to find the corresponding $P(k)$ we have used a parametrized ansatz for $P(k)$ and determined the parameters so as to reproduce the $\frac{w_{D}}{w_{T}}(w_{T})$-data as well as possible. In Fig.\ 4b we have tested three different parametrization forms. The first is a pure power law, $P_{w_T}(k)\sim 1/k^{\gamma}$, (short dashed curve in Fig.\ 4b). Our conclusion is that a power law is incompatible with the data and can be ruled out. The next try is a power law with an exponential cut off, $P_{w_T}(k)\sim \exp(-bk)/k^{\gamma}$. This form gives a very resonable approximation of the data and the function representing the binned data in Fig.\ 1a corresponds to the long dashed curve in Fig.\ 4b. But one can, off course, do a little bit better by adding another parameter. The augmentet power law with an exponential cut off, $P_{w_T}(k)\sim \exp(-bk)/(k+c)k^{\gamma-1}$, gives an even better fit to the data (open circles in Fig.\ 4b).


As simple quantitative goodness measure, one can take the maximum absolute difference between the real data and the data obtained from the various parametrizations: the values for the power-law, power-law with exponential cut off and the augmented power-law 
with exponential cut off are approximately $0.063$, $0.022$ and $0.008$, respectively. In Fig.\ 4a we have replotted the binned HE-data from Fig.\ 1a together with the best parametrization of $P(k)$ obtained from the $\frac{w_{D}}{w_{T}}(w_{T})$-data in Fig.\ 4b (circles and dashed curve, respectively). The interesting point here is that our data analysis, which makes use of the RBT-transformation, makes it possible to distinguish between parametrizations of $P(k)$ which would otherwise be very hard to distinguish. This is illustrated in Fig.\ 4c which directly compares the augemented power law with exponential cut off with the straight power law with exponential cut off. As seen from the Fig.\ 4c, there is almost no discernable difference when $P(k)$ is plotted in a log-log scale.

A consequence of the RBT-transformation is that the functional form of $P(k)$ changes with the length of the text. The full drawn curve in Fig.\ 4a gives $P(k)$ corresponding to $n=200$-parts of HE obtained from the parametrization of the form $P(k)\sim \exp(-bk)/(k+c)k^{\gamma-1}$ determined from Fig\ .4b. It agrees very well with the real data. 


In Fig.\ 3 it was demonstrated that the word frequency distribution, associated with $n$-part sections of a novel of an author, to good approximation also describes a shorter novel by the same author, provided the shorter novel has the same length as the sections. 
One can then extrapolate this idea and imagine that the longer novel also can be described as a section of an even longer novel, and so on.
This leads to the suggestion of a "meta book", a giant single "mother book" which characterizes the word-frequency distribution of all the writings of an author. An author would then, when writing a novel, be roughly pulling a section of $w_{T}$ words from this "meta book" resulting in a word-frequency distribution $P_{w_{T}}(k)$. This is the same as transforming down the "meta book" via the RBT to the size $w_{T}$. The "meta book"-concept will be further explored in a forthcoming paper.\cite{seb}

\section{Conclusions}

We have shown that the words belonging to a frequency-class in a book have a tendency to be randomly distributed thoughout the text.
This randomness is incompatible with text growth models like the Simon model\cite{simon55}. This is because these models are based on a stochastic assumption of re-using words already written in the text. This is true for all growth models, independent on the detail of the growth mechanism. It was also shown that the word-frequency distribution of a novel has a \textit{shape} which systematically depends on the size of the novel. Also this feature is incompatible the Simon model \cite{simon55}. Instead the properties of a novel were to large extent found to be shared with a random null model. The size transformation of this model is explicitly given by a Random Book Transformation (RBT) and some consequences of this were explored. 
We speculate that the word-frequency is consistent with the concept of a "meta book" which characterizes the word-frequency distribution of all the writings of an author. 

Our findings about the statistical properties of the words in a novel seem to be general: It does not matter much which author or book you pick, the overall properties are the same (at least for the English novels we have so far analyzed). Thus it does say something general about the structure of the written language used by a single author. Since language in general is a product of the human evolution, it also means that the statistical properties presumably reflects some evolutionary pressure.

\section{Acknowledgement}
This work was supported by the Swedish research Council through contract 50412501.
Very helpful discussions with Seung Ki Baek are also gratefully acknowledged.

\section{Appendix A: Collection of Books}
\begin{table}[ht!]
\caption{List of the books analyzed. $W_T$ is the total number of words in the book, $W_D$ is the total number of \emph{different} words in the book and $W_T/W_D$ is the average number of times a word is used. The initials of the authors stand for: E.M F $\rightarrow$ E.M. Forster. H M $\rightarrow$ Herman Melville. G O $\rightarrow$ George Orwell. T H $\rightarrow$ Thomas Hardy. D.H. L $\rightarrow$ D.H Lawrence.}
\begin{tabular}{l l c c c}
\hline\hline
Author & Book (abbr) & $W_T$ & $W_D$ & $W_T/W_D$ \\
\hline
E.M F & Howards End (HE) & 110.224 & 9.256 & 11,91 \\
      & The Longest Journey (LJ)& 95.265 & 8.443 & 11,28 \\
H M & White Jacket (WJ) & 143.368 & 13.710 & 10,46 \\
    & Moby Dick (MD) & 212.473 & 17.226 & 12,33 \\
G O & 1984 & 104.393 & 8.983 & 11,62 \\
T H & Jude the Obscure (JO) & 146.557 & 10.896 & 13,45 \\
D.H L & Woman in Love (WL) & 182.722 & 11.301 & 16,20 \\
      & Sons and Lovers (SL) & 162.101 & 9.606 & 16,87 \\
      & The Prussian Officer (PO) & 9.115 & 1.823 & 5.00\\
\hline
\end{tabular}
\label{book_list}
\end{table}

In order to verify the generality of our results and conclusions, a collection of eight books (in addition to Howards End) was analyzed (see table \ref{book_list}). \emph{The Prussian Officer} (PO) is not a part of the analysis in Fig.\ 2 because of its small size. It is however a part of the analysis in Fig.\ 3. 
In order to get a quantitative measure of how much the curves for the three starting points, in Fig.\ 2c and d, differ we introduce two quatities: $\xi_{rms}$ and $\xi_{\Delta}$, given by the expresions

\begin{eqnarray}
\xi_{rms} &=& \left< \sqrt{\frac{1}{W_{Ti}}\sum_{w_T=0}^{W_{Ti}} (w_{Di}-w_{Dj})^2} \right> \\
\xi_{\Delta} &=& \left< \frac{1}{W_{Ti}}\sum_{w_T=0}^{W_{Ti}} (w_{Di}-w_{Dj}) \right>
\end{eqnarray}
Where $i$ and $j$ denote the part of the book and the $\langle ... \rangle$ is an average over all the combinations of $i,j = 1,2,3$ where $i>j$.
The length of each part is $W_{Ti}=25.000$.
The first equation gives an average root mean square distance between the curves.
The second equation gives the average difference between two curves representing one part and a later part of the book. This means that if we have a trend that the curves for later parts in the book tend to have larger values for the $w_D(w_T)$-curve, then $\xi_{\Delta}$ will be a large positive number. If the trend is that later parts have smaller values we will get a large negative number. And, if there is no trend at all, we will get a value close to zero.
Figure \ref{figA1} shows the curves for the seven extra books and table \ref{book_values} shows the values of $\xi_{rms}$ and $\xi_{\Delta}$.
The \emph{Simon}-book from Fig.\ 2d and one randomized version of \emph{Howards End} (HE$_{rand}$) are also included in table \ref{book_values} to give two reference points.

\begin{table}[t!]
\caption{A list of the eight books analyzed plus the Simon-book and one randomized version of HE, showing the values for $\xi_{rms}$ and $\xi_{\Delta}$.}
\begin{tabular}{c c c c c c c c c c c}
\hline\hline
 & Simon & HE$_{rand}$ & HE & LJ & WJ & MD & 1984 & JO & WL & SL\\
\hline 
$\xi_{rms}$ & 1207 & 33 & 68 & 176 & 122 & 185 & 215 & 212 & 172 & 349\\
$\xi_{\Delta}$ & 1113 & -13 & -38 & -43 & -98 & 151 & -153 & -103 & -157 & -326\\
\hline
\end{tabular}
\label{book_values}
\end{table}

When compared to the Simon-book, all the real books seem to have small values of $\xi_{rms}$ and $\xi_{\Delta}$, indicating a strong resemblance to the null model of the random book. The values in the second row is also showing that there is no real trend among the real books, except for SL, which has a small negative trend (compared to the Simon-book which has a very strong \emph{positive} trend).

\newpage
\section{Appendix B: Simon-model}
In the Simon model a word is being written at every time step. With probability $\alpha$ a new word, that has never been written in the book so far, is written. And with probability $1-\alpha$ an old word is rewritten, chosen uniformly from the words existing in the book. This means that the probability for a word to be rewritten is proportional to the number of times it has already been written. When re-creating a real book the parameter $\alpha$ ($=\frac{W_D}{W_T}$) is usually a small number ($\sim 0.1$) and the length of the book ($T=W_T$) is generally large ($\sim 10^5$).

We want to start by calculating how big a fraction of a book, written by the Simon-model, one has to read before having encountered half of all the words that appear only once in the book.
To do this we need to calculate the probability that a specific word which is introduced at time $t$ is not repeated through out the book with length $T$.
At every time $t'$ the probability for this word not to be rewritten is the sum of the probabilities that another of the words already written is rewritten ($(1-\alpha)(\frac{t'-1}{t'})$) and that instead a completely new word is written ($\alpha$).
At time $t$, $t$ words have been written in total and $T-t$ words are still to be written, so the total probability $p(t)$ becomes

\begin{eqnarray}
p(t) &=& \prod_{t'=t}^T\left[(1-\alpha)(\frac{t'-1}{t'})+\alpha\right]\nonumber\\
&=& (1-\alpha)^{T-t}\prod_{t'=t}^T\left[1+\frac{\alpha}{1-\alpha}-\frac{1}{t'}\right]
\label{B1}
\end{eqnarray}
We introduce the quantity $\rho = 1+\frac{\alpha}{1-\alpha} = \frac{1}{1-\alpha}$ and take the logarithm on both sides of eq.\ \ref{B1}, and get

\begin{equation}
\ln p(t) = \ln \left(\frac{1}{\rho}\right)^{T-t} + \sum_{t'=t}^T\ln\left(\rho-\frac{1}{t'}\right)
\label{B2}
\end{equation}

Since $1/t' << 1$ (except for very small times, which includes only a tiny part of the whole text) we make a Taylor expansion around zero, approximate the sum with an integral and get
\begin{eqnarray}
\sum_{t'=t}^T\ln\left(\rho-\frac{1}{t'}\right) \approx \int_{t'=t}^T\left(\ln \rho - \frac{1}{t'\rho} \right)dt'\nonumber\\
= \left[t'\ln \rho - \frac{\ln t'}{\rho} \right]_t^T = \ln \rho^{T-t} + \frac{1}{\rho}\ln\left(\frac{t}{T}\right)
\label{B3}
\end{eqnarray}

Substituting Eq.\ \ref{B3} into \ref{B2} gives
\begin{eqnarray}
\ln p(t) & = & \ln \left(\frac{1}{\rho}\right)^{T-t} + \ln \rho^{T-t}+\frac{1}{\rho}\ln\left(\frac{t}{T}\right)\nonumber\\
&=& \ln\left(\frac{t}{T}\right)^{\frac{1}{\rho}} \Rightarrow \underline{p(t) = \left(\frac{t}{T}\right)^{1-\alpha}}
\end{eqnarray}

If we write a book, then $p(t)$ is the average number of $k=1$-words one gets from the introduction time $t$, and so
\begin{equation}
\sum_{t=1}^T p(t) = W_D(1),
\end{equation}
where $W_D(1)$ is the total number of $k=1$-words in the book.

\begin{eqnarray}
W_D(1) &=& \sum_{t=1}^T \left(\frac{t}{T}\right)^{1-\alpha} = 
\left\{\textrm{substituting}\ \frac{t}{T} = x\right\}\nonumber\\
&\approx& T\int_{1/T}^1 x^{1-\alpha}dx = T\left[\frac{x^{2-\alpha}}{2-\alpha} \right]_{1/T}^1\nonumber\\ 
&=& \frac{T}{2-\alpha}\left(1-\underbrace{\left(\frac{1}{T}\right)^{2-\alpha}}_{\approx 0} \right)\nonumber\\
\Rightarrow W_D(1) &\approx& \frac{T}{2-\alpha}
\end{eqnarray}

To find the time, $T_{1/2}$, when we have introduced half of all the $k=1$-words, we solve the expression:
\begin{eqnarray}
&& \sum_{t=1}^{T_{1/2}} p(t) = \frac{W_D(1)}{2}\\
\Rightarrow && \frac{1}{W_D(1)}\sum_{t=1}^{T_{1/2}} p(t) = \frac{1}{2}\nonumber\\
&& \frac{1}{2} = \frac{2-\alpha}{T}\sum_{t=1}^{T_{1/2}}\left(\frac{t}{T}\right)^{1-\alpha}
= \left\{\textrm{substituting}\ \frac{t}{T} = x\right\}\nonumber\\
&& \approx (2-\alpha)\int_{1/T}^{T_{1/2}/T}x^{1-\alpha}dx = (2-\alpha)\left[\frac{x^{2-\alpha}}{2-\alpha} \right]_{1/T}^{T_{1/2}/T}\nonumber\\
&& = \left(\frac{T_{1/2}}{T} \right)^{2-\alpha}-\underbrace{\left(\frac{1}{T} \right)^{2-\alpha}}_{\approx 0} \approx \left(\frac{T_{1/2}}{T} \right)^{2-\alpha}\nonumber\\
\Rightarrow && \frac{T_{1/2}}{T} = \left(\frac{1}{2}\right)^{\frac{1}{2-\alpha}}
\label{T_1/2}
\end{eqnarray}
Which is the fraction of the book one has to read before one half of the $k=1$-words have been read.
For the Simon-book in Fig.\ 2 ($\alpha = 0.083$) this value is $\frac{T_{1/2}}{T}=0.697$. That is, $69.7 \%$ of the book. \\

Equation \ref{T_1/2} can be generalized into
\begin{equation}
\frac{T_n}{T} = n^{\frac{1}{2-\alpha}}
\end{equation}
Where $n$ is the fraction of one-degree words.\\

Next we want to do the same thing for $k=2$-words. 
Now we need to calculate the probability that if a word is first introduced at time $t_1$ it will only be repeated once at time $t_2$.
This probability is given by

\begin{equation}
p(t_1,t_2) = \prod_{t'=t_1}^{t_2}\left[\rho - \frac{1}{t'}\right]\frac{1}{\rho}\left(\frac{1}{t^2}\right) \prod_{t'=t_2}^T\left[\rho - \frac{2}{t'}\right].
\end{equation}
where the $2$ in the last product comes from now having two words with the possibility of being picked.

This equation can be evaluated in a similary way as for the $k=1$-case, and we get:
\begin{equation}
p(t_1,t_2) = T^{2(\alpha-1)}t_1^{1-\alpha}t_2^{-\alpha}
\end{equation}

Again, this quantity gives the average number of $k=2$-words one will get from words that are introduced at time $t_1$ and repeated at time $t_2$, which means that
\begin{equation}
\sum_{t_1=1}^T\sum_{t_2=t_1}^T p(t_1,t_2) = W_D(2)
\end{equation}
where we sum over all possible combinations of $t_1$ and $t_2$ where $t_2 > t_1$.
This can also be evaluated in a similar way as for the $k=1$-case and we get

\begin{equation}
W_D(2) \approx \frac{T}{1-\alpha}\left(\frac{1}{2-\alpha} - \frac{1}{3-2\alpha}\right).
\end{equation}

The \emph{total} number of words in a $k$-group (all the destinct words with frequency $k$) is $kW_D(k)$. The time $T_{1/2}$, when we have read half of all these words, is given by the expresion

\begin{eqnarray}
2\sum_{t_1=1}^{T_{1/2}}\sum_{t_2=t_1}^{T_{1/2}} p(t_1,t_2)+\sum_{t_1=1}^{T_{1/2}}\sum_{t_2=T_{1/2}}^T p(t_1,t_2) = \frac{2W_D(2)}{2}
\label{half_k2}
\end{eqnarray}
The first sum counts all the words where both its appearances happen before $T_{1/2}$ and is thus counted twice. The second sum counts all the words that was introduced before $T_{1/2}$ and repeated after $T_{1/2}$ and is thus counted as one. Equation \ref{half_k2} can be evaluated into:
\begin{equation}
\frac{\left(\frac{T_{1/2}}{T}\right)^{3-2\alpha}\left(\frac{1}{2-\alpha}-\frac{2}{3-2\alpha}\right)+\frac{1}{2-\alpha}\left(\frac{T_{1/2}}{T}\right)^{2-\alpha}}{\left(\frac{1}{2-\alpha}-\frac{1}{3-2\alpha}\right)} = 1
\label{T_1/2_k2}
\end{equation}
Equation \ref{T_1/2_k2} cannot be solved analytically but a numerical solution for the Simon-book in Fig.\ 2 ($\alpha = 0.083$) gives the value $\frac{T_{1/2}}{T} \approx 0.638$.

We now have two points ($k=1$ and $k=2$) giving the asymptotic functional form for low k:s.
In Fig.\ 2b a straight line was drawn intersecting these two point ($\frac{T_{1/2}}{T}_{k=1}=0.697$ and $\frac{T_{1/2}}{T}_{k=2}=0.638$) to show this asymptotic behavior.

The derivations for this quantity gets very complicated for larger values of $k$ since we are summing over all different words with the same frequency. But for very large $k$:s we have words that are alone in their frequency-group. That is, they are the only one with that particularly frequency. This makes the derivation much simpler and we can get the asymptotic behavior for large $k$:s.
From Ref.\ \cite{barabasi99} we get the equation

\begin{equation}
k(t) = \left(\frac{T}{t}\right)^{1-\alpha}
\end{equation}
where $k(t)$ is the number of occurrences a word will have in a book of length $T$ if it was introduced at time $t$.
We want to know at what time we have written half of those words. This is given by

\begin{eqnarray}
&&k_{1/2}(t) = \frac{k(t)}{2} = \left(\frac{T_{1/2}}{t}\right)^{1-\alpha}\nonumber\\
\Rightarrow && \frac{k(t)}{k_{1/2}(t)} = 2 = \frac{\left(\frac{T}{t}\right)^{1-\alpha}}{\left(\frac{T_{1/2}}{t}\right)^{1-\alpha}}
= \left(\frac{T_{1/2}}{T}\right)^{-(1-\alpha)}\nonumber\\
\Rightarrow && \frac{T_{1/2}}{T} = 2^{-\frac{1}{1-\alpha}}
\end{eqnarray}
This equation holds for all k-values where $W_D(k)=1$. For the Simon-book in Fig.\ 2 ($\alpha = 0.083$)
This value is $\frac{T_{1/2}}{T} \approx 0.47$ and represents the horizontal line if Fig.\ 2b.

\newpage FIGURE CAPTIONS\newline\newline

Fig 1: Word frequency distribution $P(k)$ for the book Howards End (HE): a)
Circles give the raw data. The horizontal tail reflects that the largest
number of occurrences corresponds to single words. Triangles give log-binned
data and follow a smooth curve implying a stochastic origin. The actual data
is to good approximation of the form $P(k)\sim\exp(-bk)/k^{\gamma}$ with
$\gamma=1.73$: b) $P(k)$ changes with the section size of the book. Full curve
represent the complete HE, long-dashed curve and short-dashed curves represents sections
corresponding to a 20th and 200th parts of HE, respectively. The curves
represent the log-binned data.
\newline\newline

Fig 2: Number of distinct words $w_{D}(w_{T})$ as a function of the total
number of words $w_{T}$: a) Real and randomized HE given by full and dashed
curve, respectively. The close agreement implies that the words are close to randomly distributed throughout the book: 
b) Curves describing how big a fraction of the book one has to read before having encountered
half of all the words with a specific frequancy.
The circles and triangles represent the real HE and a Simon-book (same size and $\langle k\rangle$ as HE)
respectively. The dashed lines are showing the analytic asymptotic behavior of the Simon-book (see appendix B).
The full line represents the average result for a randomized book and the gray areas shows
one and two standard deviations away from the random book.
c) $w_{D}(w_{T})$ for three different starting points within the book; full, long-dashed and
short-dashed curves correspond to the beginning, middle and end of HE,
respectively. The close agreement implies that the word distribution in a book
is to good approximation translational invariant: d) The same different
starting points as in c) assuming that the word-distribution was given by the
Simon text growth model. The large and systematic differences shows that the
Simon-type growth models do not describe the randomness of the word
distribution in a real text.
\newline\newline

Fig 3: The sectioning of two full novels compared to a short story by the same author.
a) The circles represent the binned data of the full novel \emph{Woman in Love}. The triangles
show the sectioning (a 20th-part) of the same book down to the same size as the short story \emph{the Prussian Officer}, shown with squares.
b) The same as for a) but for the full noval \emph{Sons and Lovers} sectioned into an 18th-part.
\newline\newline

Fig 4: The random book transformation (RBT). a) the data for HE (open circles)
is parametrized (dashed curve). The dashed curve is transformed to a
200th-part of the book (full curve). This full curve should correspond to a
200th-part of the randomized HE (open triangles). The agreement is striking.
The distribution corresponding to a 200th part of the real HE is given by the
open squares. The close agreement with the triangles shows that the words are
to large extent randomly distributed. b) The function $\frac{w_D}{w_T}(w_T)$ 
for HE: Full curve corresponds to the randomized HE and the circles
are obtained from the RBT using the parametrization of $P(k)$ given in a). The
agreement is perfect. The long-dashed curve corresponds to the data obtained
from RBT using the parametrization of $P(k)$ given in Fig.\ 1a and the inset, which is an 
in-zoomed version of the dashed squar, is showing how this curve is deviating from the real data.
The short dashed curve in b) represents a power-law fit to the word-frequancy 
distribution which clearly fails to represent the data. c) is showing how similar 
the two parametrizations are which means that RBT determines $P(k)$ to high accuracy.
\newline\newline

Fig A1: Complementary figure to Fig.\ 2a and c showing the number of distinct words $w_{D}(w_{T})$ as a function of the total
number of words $w_{T}$ for seven additional books: First column represents counting from start to finish
and the second column represents counting through three consecutive parts of the same size.

\newpage
\begin{figure}[th]
\begin{center}
\includegraphics[width=0.7\columnwidth]{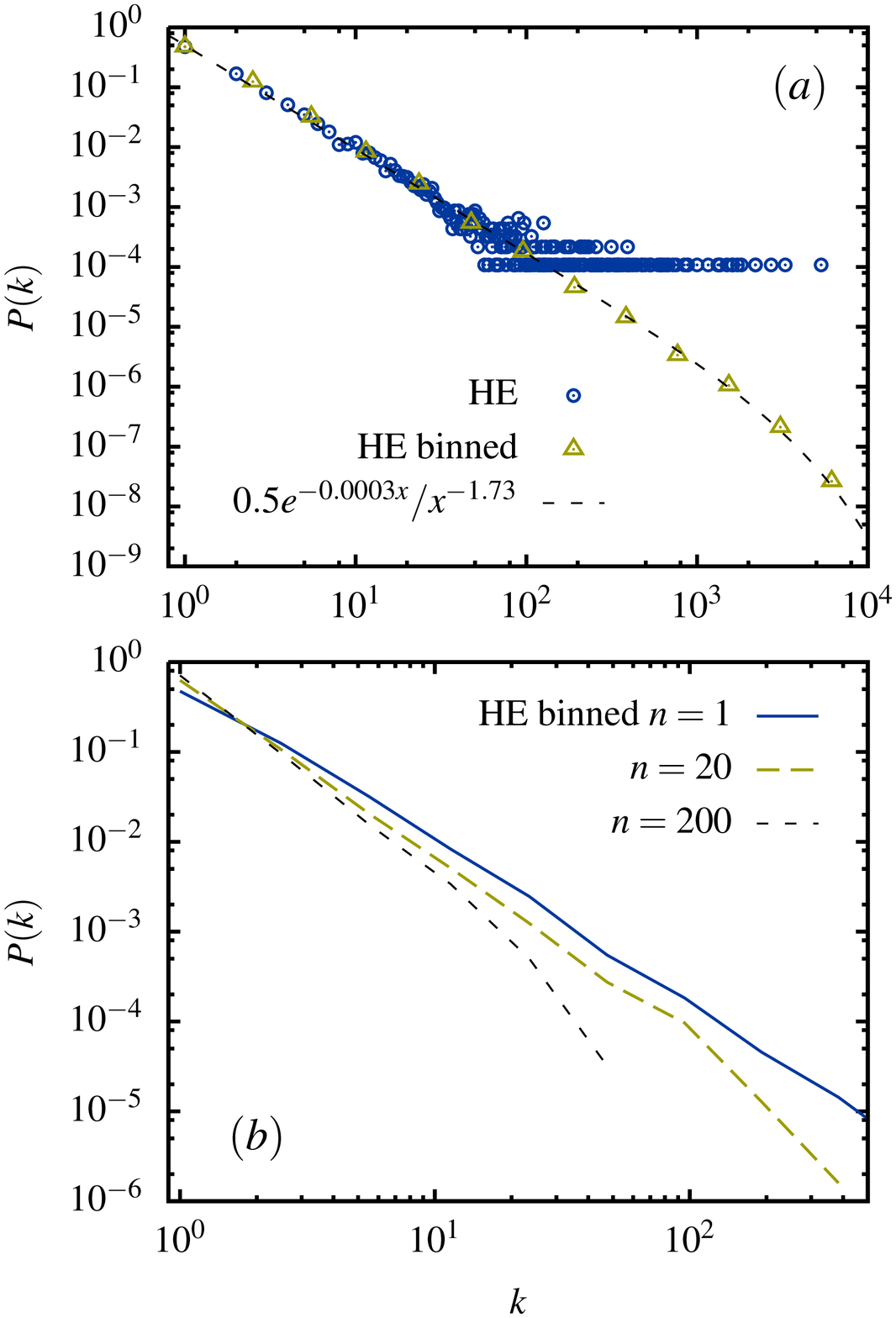}
\label{HowardsEnd}
\end{center}
\end{figure}

\begin{figure}[ptb]
\begin{center}
\includegraphics[width=0.9\columnwidth]{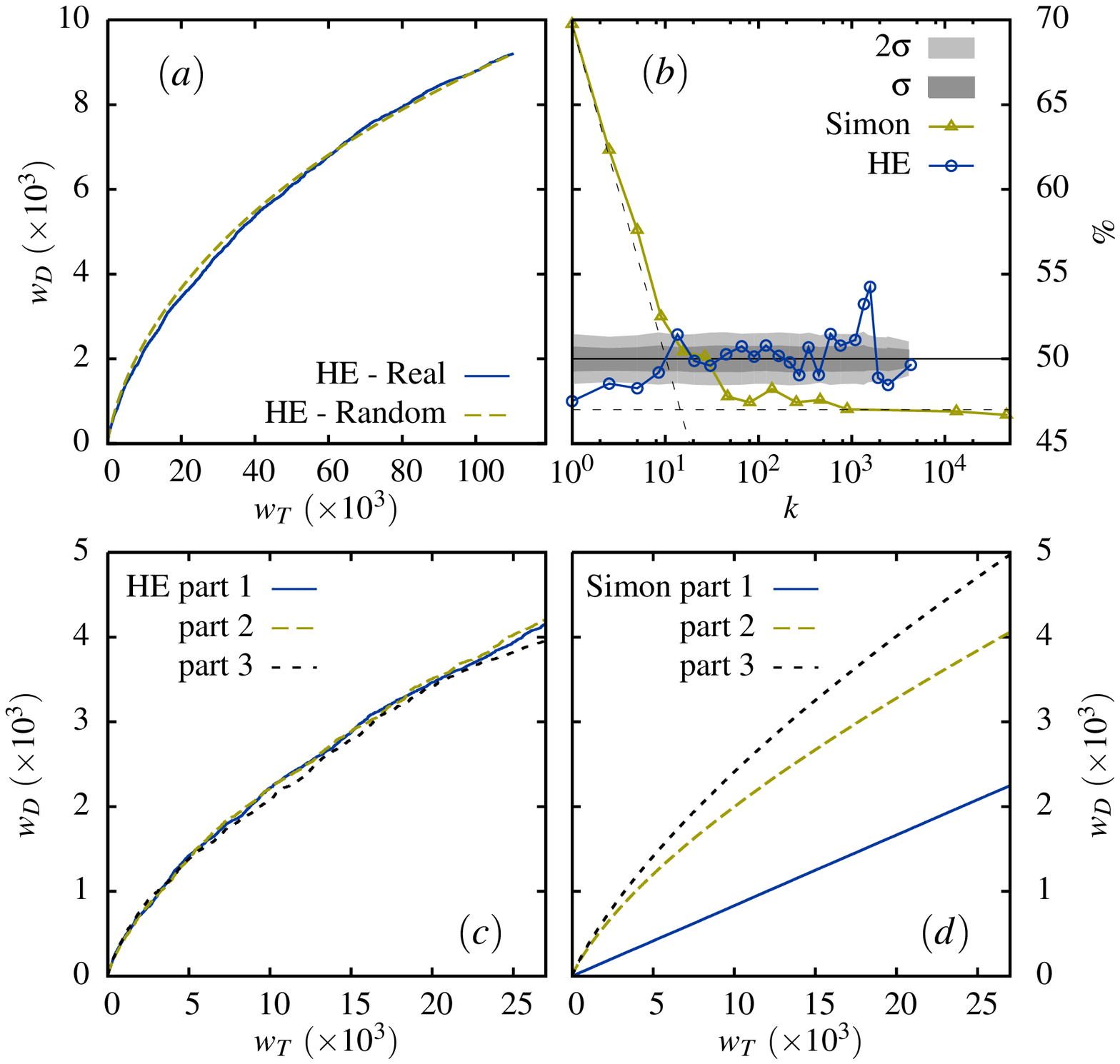}
\end{center}
\end{figure}

\begin{figure}[ptb]
\begin{center}
\includegraphics[width=0.7\columnwidth]{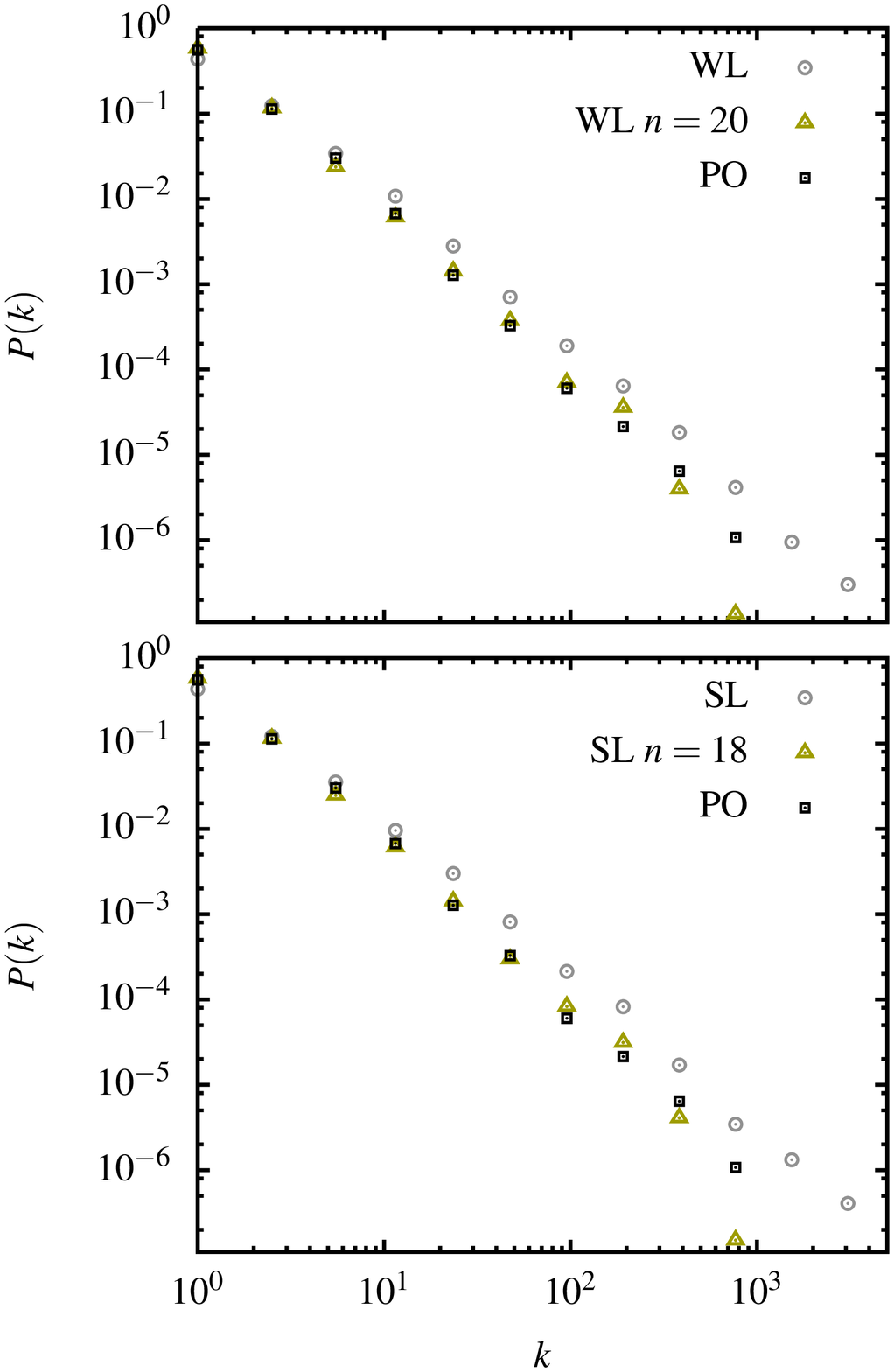}
\end{center}
\end{figure}

\begin{figure}[ptb]
\begin{center}
\includegraphics[width=0.55\columnwidth]{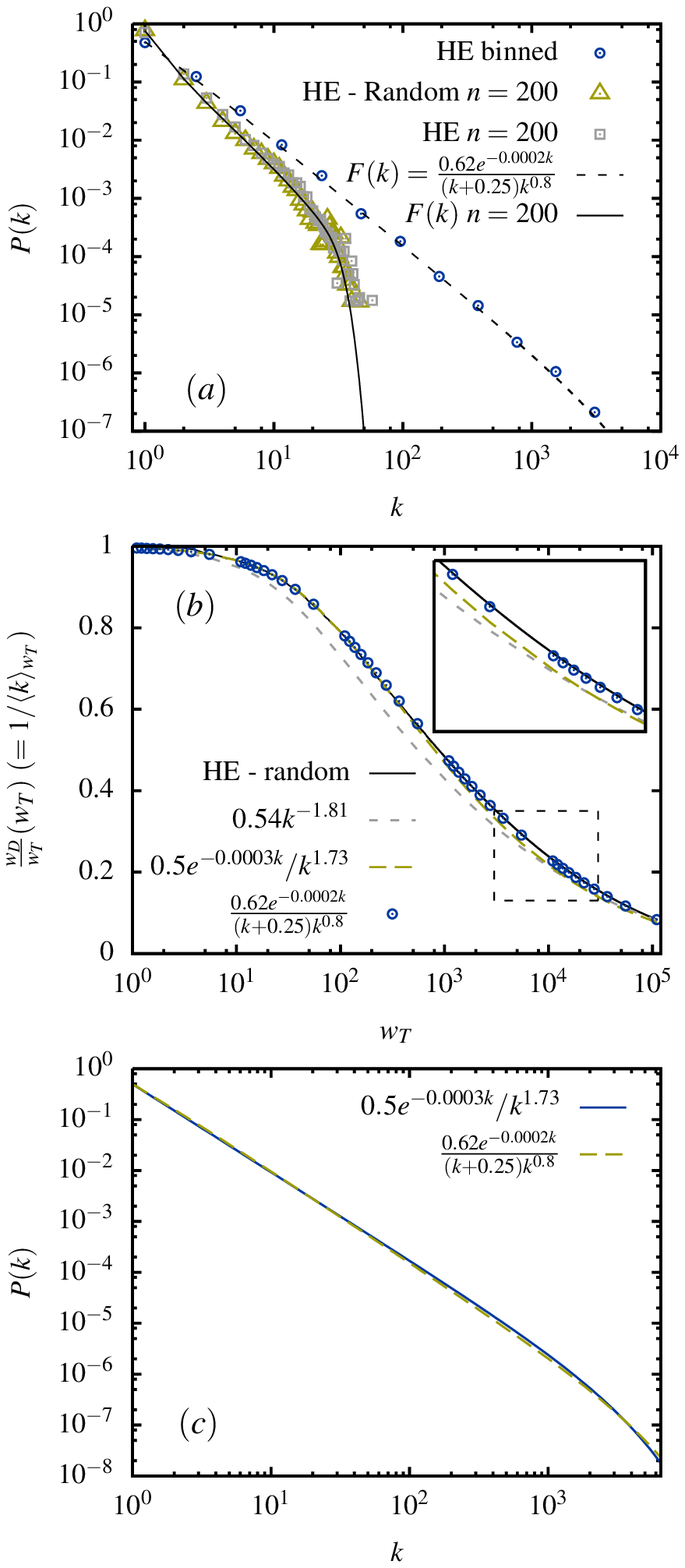}
\end{center}
\end{figure}

\begin{figure}[ptb]
\begin{center}
\includegraphics[width=0.5\columnwidth]{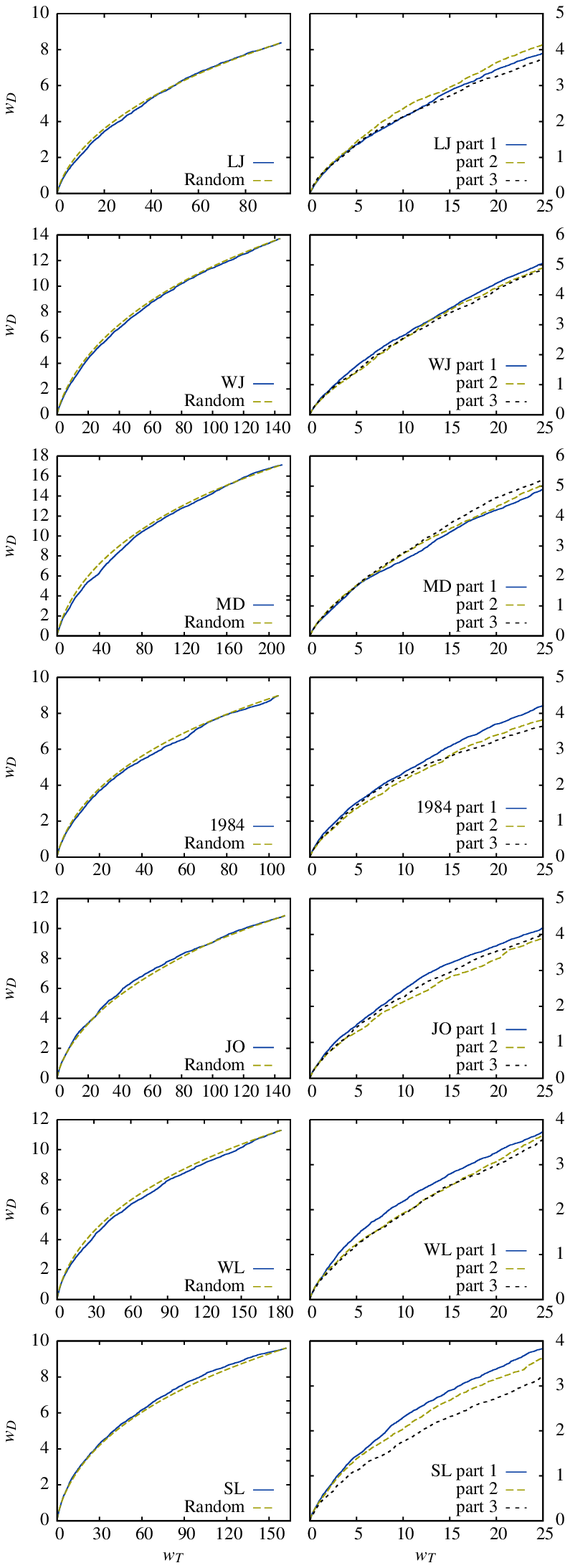}
\end{center}
\label{figA1}
\end{figure}
\end{document}